\documentclass[11pt,a4paper]{article}
\usepackage[hyperref]{naaclhlt2018}
\usepackage{times}
\usepackage{latexsym}

\usepackage{amsfonts}
\usepackage{amsmath}

\usepackage{paralist}
\usepackage{booktabs} 
\usepackage{hyperref}

\usepackage{graphicx}
\usepackage{tabularx}
\usepackage{tabulary}
\usepackage{todonotes}
\usepackage{multirow}
\usepackage{enumitem}
\usepackage{url}
\hypersetup{
  pdftitle = {Learning to Generate Wikipedia Summaries for Underserved Languages from Wikidata}
}

\newcolumntype{L}[1]{>{\raggedright\let\newline\\\arraybackslash\hspace{0pt}}m{#1}}
\newcolumntype{C}[1]{>{\centering\let\newline\\\arraybackslash\hspace{0pt}}m{#1}}
\newcolumntype{R}[1]{>{\raggedleft\let\newline\\\arraybackslash\hspace{0pt}}m{#1}}

\aclfinalcopy 

\title{Learning to Generate Wikipedia Summaries for Underserved Languages from Wikidata}

\author{Lucie-Aim\'{e}e Kaffee$^{\mathbf{1}\mathbf{\dagger}}$ \quad Hady Elsahar$^{\mathbf{2}\mathbf{\dagger}}$ \quad Pavlos Vougiouklis$^{\mathbf{1}\mathbf{\dagger}}$ \\
\textbf{Christophe Gravier}$^{\mathbf{2}}$ \quad \textbf{Fr\'{e}d\'{e}rique Laforest}$^{\mathbf{2}}$ \quad \textbf{Jonathon Hare}$^{\mathbf{1}}$ \quad \textbf{Elena Simperl}$^{\mathbf{1}}$\\\\
 $^{\mathbf{1}}$  School of Electronics and Computer Science, University of Southampton, UK  \\
 \small{\texttt{\{kaffee, pv1e13, jsh2, e.simperl\}@ecs.soton.ac.uk}}\\
  $^{\mathbf{2}}$ Laboratoire Hubert Curien, CNRS, UJM-Saint-\'{E}tienne, Universit\'{e} de Lyon, France  \\
  \small{\texttt{\{hady.elsahar, christophe.gravier, frederique.laforest\}@univ-st-etienne.fr}}}
  
\begin{document}
\maketitle

\begin{abstract}
While Wikipedia exists in $287$ languages, its content is unevenly distributed among them.
In this work, we investigate the generation of open domain Wikipedia summaries in underserved languages using structured data from Wikidata.
To this end, we propose a neural network architecture equipped with copy actions that learns to generate single-sentence and comprehensible textual summaries from Wikidata triples.
We demonstrate the effectiveness of the proposed approach by evaluating it against a set of baselines on two languages of different natures: Arabic, a morphological rich language with a larger vocabulary than English, and Esperanto, a constructed language known for its easy acquisition.
\end{abstract}

\section{Introduction}\label{sec:intro}
\renewcommand{\thefootnote}{\fnsymbol{footnote}}
\setcounter{footnote}{2}
\footnotetext{The authors contributed equally to this work.}
\renewcommand{\thefootnote}{\arabic{footnote}}
\setcounter{footnote}{0}

Despite the fact that Wikipedia exists in $287$ languages, the existing content is unevenly distributed.
The content of the most under-resourced Wikipedias is maintained by a limited number of editors -- they cannot curate the same volume of articles as the editors of large Wikipedia language-specific communities. 
It is therefore of the utmost social and cultural interests to address languages for which native speakers have only access to an impoverished Wikipedia. 
In this paper, we propose an automatic approach to generate textual summaries that can be used as a starting point for the editors of the involved Wikipedias.
We propose an end-to-end trainable model that generates a textual summary given a set of KB triples as input.
%
We apply our model on two languages that have a severe lack of both editors and articles on Wikipedia: 
Esperanto is an easily acquired artificially created language which makes it less data needy and a more suitable starting point for exploring the challenges of this task. 
Arabic is a morphologically rich language that is much more challenging to work, mainly due to its significantly larger vocabulary.
As shown in Table~\ref{tab:wikistats} both Arabic and Esperanto suffer a severe lack of content and active editors compared to the English Wikipedia which is currently the biggest one in terms of number of articles.

%
Our research is mostly related to previous work on adapting the general encoder-decoder framework for the generation of Wikipedia summaries \cite{lebret2016,Chisholm2017,DBLP:journals/corr/abs-1711-00155}.
Nonetheless, all these approaches focus on task of biographies generation, and only in English -- the language with the most language resources and knowledge bases available.
In contrast with these works, we explore the generation of sentences in an open-domain, multilingual context.
The model from \cite{lebret2016}  takes the Wikipedia infobox as an input, while \cite{Chisholm2017} uses a sequence of slot-value pairs extracted from Wikidata. Both models are only able to generate single-subject relationships.
In our model the input triples go beyond the single-subject relationships of a Wikipedia infobox or a Wikidata page about a specific item (Section~\ref{sec:model}).
Similarly to our approach, the model proposed by \cite{DBLP:journals/corr/abs-1711-00155} accepts a set of triples as input, however, it leverages instance-type-related information from DBpedia in order to generate text that addresses rare or unseen entities. 
Our solution is much broader since it does not rely on the assumption that unseen triples will adopt the same pattern of properties and entities' instance types pairs as the ones that have been used for training.
To this end, we use copy actions over the labels of entities in the input triples. 
This relates to previous works in machine translation which deals with rare or unseen word problem for translating names and numbers in text. 
\cite{Luong15} propose a model that generates positional placeholders pointing to some words in source sentence and copy it to target sentence (\textit{copy actions}).
\cite{gulcehre2016noisy} introduce separate trainable modules for copy actions to adapt to highly variable input sequences, for text summarisation. 
For text generation from tables, \cite{lebret2016} extend positional copy actions to copy values from fields in the given table.
For Question Generation,~\cite{Serban16} use a placeholder for the subject entity in the question to generalise to unseen entities. 

We evaluate our approach by measuring how close our synthesised summaries can be to actual summaries in Wikipedia  against two other baselines of different natures: a language model, and an information retrieval template-based solution. 
Our model substantially outperforms all the baselines in all evaluation metrics in both Esperanto and Arabic.
%
In this work we present the following contributions: i) We investigate the task of generating textual summaries from Wikidata triples in underserved Wikipedia languages across multiple domains, and ii) We use an end-to-end model with copy actions adapted to this task.
Our datasets, results, and experiments are available at:~\url{https://github.com/pvougiou/Wikidata2Wikipedia}.
\begin{table}[t]
  \small
  \centering
  \begin{tabularx}{0.45\textwidth}{lllll}\toprule
    \textbf{} & \textbf{Arabic} & \textbf{Esperanto} & \textbf{English} \\\midrule
    \# of Articles & 541,166 & 241,901 & 5,483,928 \\
    \# of Active Users & 7,818 & 2,849 & 129,237 \\
    Vocab. Size & 2.2M & 1.5M & 2.0M \\
    \bottomrule
  \end{tabularx}
  \caption{\label{tab:wikistats}Recent page statistics and total number of unique words (vocab. size) of Esperanto, Arabic and English Wikipedias.}
\end{table}

\begin{table*}
  \begin{center}
    \footnotesize
    \begin{tabular}{llll}\toprule
      \multirow{3}{*}{\textbf{Triples}} & \texttt{Q490900} (Floridia) & \texttt{P31} (estas) & \texttt{Q747074} (komunumo de Italio) \\
      & \texttt{Q490900} (Floridia) & \texttt{P17} (\^{s}tato) & \texttt{Q38} (Italio) \\
      & \texttt{Q30025755} (Floridia) & \texttt{P1376} (\^{c}efurbo de) & \texttt{Q490900} (Floridia) \\ \midrule
      
      \textbf{Textual Summary} & \multicolumn{3}{l}{Floridia estas komunumo de Italio.} \\\midrule
      \textbf{Vocab. Extended} & \multicolumn{3}{l}{\texttt{[[Q490900, Floridia]]} estas komunumo de \texttt{[[P17]]}.} \\ \bottomrule
    \end{tabular}
    \caption{\small{Training example: a set of triples about \textit{Floridia}. Subsequently, our system summarises the input set in the form of text. The vocabulary extended summary is the one on which we train our model.}}
    \label{table:SimpleTextGenerationExample}
  \end{center}
\end{table*}

\section{Model}\label{sec:model}
Our approach is inspired by similar encoder-decoder architectures that have already been employed on similar text generative tasks \cite{Serban16,DBLP:journals/corr/abs-1711-00155}.

\subsection{Encoding the Triples}
The encoder part of the model is a feed-forward architecture that encodes the set of input triples into a fixed dimensionality vector, which is subsequently used to initialise the decoder.
Given a set of un-ordered triples $F_E = \{ f_1,f_2, \ldots, f_R: f_j=(s_j,p_j,o_j)\}$, where $s_j$, $p_j$ and $o_j$ are the one-hot vector representations of the respective subject, property and object of the $j$-th triple, we compute an embedding $h_{f_j}$ for the $j$-th triple by forward propagating as follows:
\begin{align}
  h_{f_j} & = q(\mathbf{W_{h}}[\mathbf{W_{in}}s_j;\mathbf{W_{in}}p_j;\mathbf{W_{in}}o_j])
  \enspace, \\
  h_{F_E} & = \mathbf{W_{F}}[h_{f_1}; \ldots; h_{f_{R-1}}; h_{f_R}] \enspace,
\end{align} 
where 
$h_{f_j}$ is the embedding vector of each triple $f_j$,
$h_{F_E}$ is a fixed-length vector representation for all the input triples $F_E$. 
$q$ is a non-linear activation function, $[\ldots;\ldots]$ represents vector concatenation. $\mathbf{W_{in}}$,$\mathbf{W_{h}}$,$\mathbf{W_{F}}$ are trainable weight matrices. Unlike \cite{Chisholm2017}, our encoder is agnostic with respect to the order of input triples. 
As a result, the order of a particular triple $f_j$ in the triples set does not change its significance towards the computation of the vector representation of the whole triples set, $h_{F_E}$.

\subsection{Decoding the Summary}
\begin{figure}[t]
  \centering
  \def\svgwidth{.99\linewidth}
  \includegraphics[width=0.985\linewidth]{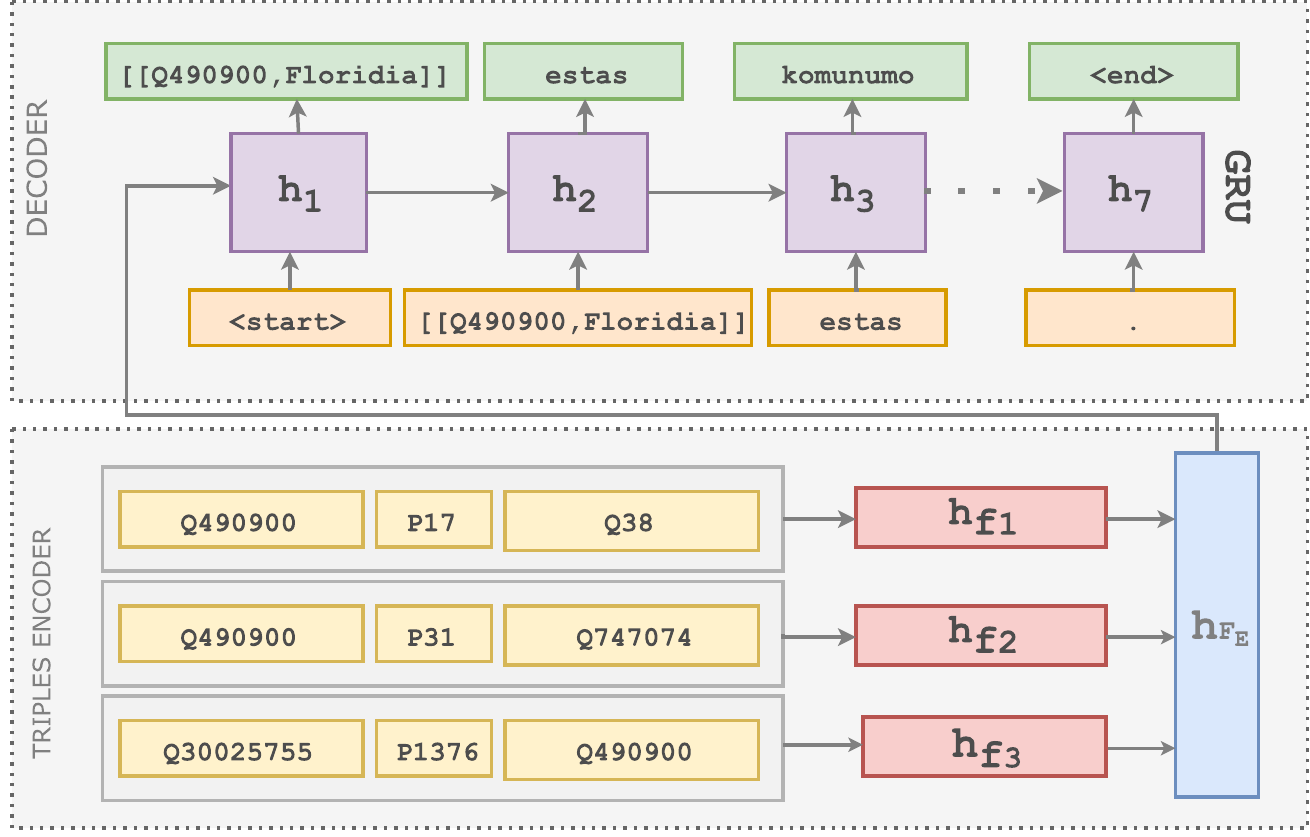}
  \caption{Model Overview}
  \label{fig:MultilingualArchitecture}
\end{figure}

The decoder part of the architecture is a multi-layer RNN~\cite{Cho2014} with Gated Recurrent Units which generates the textual summary one token at a time.
The hidden unit of the GRU at the first layer is initialised with $h_{F_E}$. 
At each timestep $t$, the hidden state of the GRU is calculated as follows:
\begin{align} 
  h_t^l = \text{GRU}(h_{t-1}^l, h_{t}^{l-1})
\end{align}
The conditional probability distribution over each token $y_t$ of the summary at each timestep $t$ is computed as the $\text{softmax}(\mathbf{W_{out}}h_t^L)$ over all the possible entries in the  summaries dictionary, where $h_t^L$ is the hidden state of the last layer and $\mathbf{W_{out}}$ is a biased trainable weight matrix. \\
A summary consists of words and mentions of entity in the text. We adapt the concept of \textit{surface form tuples} \cite{DBLP:journals/corr/abs-1711-00155} in order to be able to learn an arbitrary number of different lexicalisations of the same entity in the summary (e.g. ``aktorino'', ``aktoro''). 
Figure \ref{fig:MultilingualArchitecture} shows the architecture of our generative model when it is provided with the three triples of the idealised example of Table \ref{table:SimpleTextGenerationExample}.

\subsection{Copy Actions}\label{sec:vocab-ext}
Following~\cite{Luong15,lebret2016} we model all the copy actions on the data level through a set of special tokens added to the basic vocabulary. 
Rare entities identified in text and existing in the input triples are being replaced by the token of the property of the relationship to which it was matched. We refer to those tokens as \textit{property placeholders}.
In Table~\ref{table:SimpleTextGenerationExample}, \texttt{[[P17]]} in the vocabulary extended summary is an example of property placeholder -- would it be generated by our model, it is replaced with the label of the object of the triple with which they share the same property (i.e. \texttt{Q490900} (Floridia) \texttt{P17} (\^{s}tato) \texttt{Q38} (Italio)).
When all the tokens of the summary are sampled, each property placeholder that is generated is mapped to the triple with which it shares the same property and is subsequently replaced with the textual label of the entity. 
We randomly choose an entity, in case there are more than one triple with the same property in the input triples set.
%

\subsection{Implementation and Training Details}
We implemented our neural network models using the Torch\footnote{\href{http://torch.ch}{Torch} is a scientific computing package for Lua. It is based on the \href{http://luajit.org/}{LuaJIT} package.} package.

We included the 15,000 and 25,000 most frequent tokens (i.e. either words or entities) of the summaries in Esperanto and Arabic respectively for target vocabulary of the textual summaries.
Using a larger size of target dictionary in Arabic is due to its greater linguistic variability -- Arabic vocabulary is $47\%$ larger than Esperanto vocabulary (cf. Table~\ref{tab:wikistats}).
We replaced any rare entities in the text that participate in relations in the aligned triples set with the corresponding property placeholder of the upheld relations.
We include all property placeholders that occur at least $20$ times in each training dataset. 
Subsequently, the dictionaries of the Esperanto and Arabic summaries are expanded by $80$ and $113$ property placeholders respectively. 
In case the rare entity is not matched to any subject or object of the set of corresponding triples it is replaced by the special \texttt{<resource>} token.
Each summary is augmented with the respect start-of-summary \texttt{<start>} and end-of-summary \texttt{<end>} tokens.

For the decoder, we use $1$ layer of GRUs. We set the dimensionality of the decoder's hidden state to $500$ in Esperanto and $700$ in Arabic.
We initialise all parameters with random uniform distribution between $-0.001$ and $0.001$, and we use Batch Normalisation before each non-linear activation function and after each fully-connected layer \cite{Ioffe2015} on the encoder side \cite{DBLP:journals/corr/abs-1711-00155}.
During training, the model tries to learn those parameters that minimise the sum of the negative log-likelihoods of a set of predicted summaries. The networks are trained using mini-batch of size $85$. 
The weights are updated using Adam~\cite{Kingma2014} (i.e. it was found to work better than Stochastic Gradient Descent, RMSProp and AdaGrad) with a learning rate of $10^{-5}$. 
An $l_2$ regularisation term of $0.1$ over each network's parameters is also included in the cost function.

The networks converge after the $9$th epoch in the Esperanto case and after the $11$th in the Arabic case.
During  evaluation and testing, we do beam search with a beam size of $20$, and we retain only the summary with the highest probability.
We found that increasing the beam size resulted not only in minor improvements in terms of performance but also in a greater number of fully-completed generated summaries (i.e. summaries for which the special end-of-summary \texttt{<end>} token is generated). 
%

\section{Dataset}\label{sec:dataset}
In order to train our models to generate summaries from Wikidata triples, we introduce a new dataset for text generation from KB triples in a multilingual setting and align it with the triples of its corresponding Wikidata Item. 
For each Wikipedia article, we extract and tokenise the first introductory sentence and align it with triples where its corresponding item appears as a subject or an object in the Wikidata truthy dump. 
In order to create the \textit{surface form tuples} (i.e. Section~\ref{sec:vocab-ext}), we identify occurrences of entities in the text along with their verbalisations. 
We rely on keyword matching against labels from Wikidata expanded by the global language fallback chain introduced by Wikimedia\footnote{\url{https://meta.wikimedia.org/wiki/Wikidata/Notes/Language_fallback}} to overcome the lack of non-English labels in Wikidata~\cite{kaffee2017glimpse}.

For the \textit{property placeholders}, we use the distant supervision assumption for relation extraction~\cite{mintz_2009}. %
\begin{table}[t]
\centering
\small
\setlength{\extrarowheight}{1.5pt}
\begin{tabular}{|L{2.8cm}|cc|}
\hline
& \textbf{Arabic} & \textbf{Esperanto} \\ \hline
Avg. \# of Tokens per Summary  & $28.1$ ($\pm 28.8$) & $26.4$ ($\pm 22.7$) \\ \hline
Avg. \# of Triples per Summary & $8.1$ ($\pm 11.2$) & $11.0$ ($\pm 13.8$) \\ \hline
Avg. \# of Linked Named Entities & $2.2$ ($\pm 1.0$) & $2.4$ ($\pm 1.1$) \\ \hline
Avg. \# of Aligned Triples & $0.1$ ($\pm 0.4$) & $0.2$ ($\pm 0.5$) \\ \hline\hline
Vocabulary Size & $344,827$ & $226,447$ \\
Total \# of Summaries & $255,741$ & $126,714$ \\
\hline
\end{tabular}
\caption{\label{table:ArabicEsperantoDatasetsStatistics}Dataset statistics in Arabic and Esperanto.}
\end{table}
Entities that participate in relations with the main entity of the article are being replaced with their corresponding property placeholder tag.
Table \ref{table:ArabicEsperantoDatasetsStatistics} shows statistics on the two corpora that we used for the training of our systems.

\section{Baselines}
To demonstrate the effectiveness of our approach, we compare it to two competitive systems.
\begin{table*}[!h]
\small
\centering
\setlength{\extrarowheight}{1.5pt}
\begin{tabular}{@{}|l|l|cc|cc|cc|cc|cc|cc|@{}}
  \hline
  & \multirow{2}{*}{\textbf{Model}} & \multicolumn{2}{c|}{\textbf{BLEU 1}} & \multicolumn{2}{c|}{\textbf{BLEU 2}} & \multicolumn{2}{c|}{\textbf{BLEU 3}} & \multicolumn{2}{c|}{\textbf{BLEU 4}} & \multicolumn{2}{c|}{\textbf{ROUGE\textsubscript{L}}} & \multicolumn{2}{c|}{\textbf{METEOR}} \\
  & & Valid. & Test & Valid. & Test & Valid. & Test & Valid. & Test & Valid & Test & Valid. & Test \\ 
  \hline
  \multirow{8}{*}{\raisebox{\normalbaselineskip}[0pt][0pt]{\rotatebox[origin=c]{90}{\textbf{\footnotesize{Arabic}}}}}    
  & KN & 12.84 & 12.85 & 2.28 & 2.4 & 0.95 & 1.04 & 0.54 & 0.61 & 17.08 & 17.09 & 29.04 & 29.02 \\
  & KN\textsubscript{ext} & 28.93 & 28.84 & 21.21 & 21.16 & 16.78 & 16.76 & 13.42 & 13.42 & 28.57 & 28.52 & 30.47 & 30.43 \\
  & IR & 41.39 & 41.73 & 34.18 & 34.58 & 29.36 & 29.72 & 25.68 & 25.98 & 43.26 & 43.58 & 32.99 & 33.33 \\
  & IR\textsubscript{ext} & 49.87 & 48.96 & 42.44 & 41.5 & 37.29 & 36.41 & 33.27 & 32.51 & 51.66 & 50.57 & 34.39 & 34.25 \\
  & Ours & 53.61 & 54.26 & 47.38 & 48.05 & 42.65 & 43.32 & 38.52 & 39.20 & 64.27 & 64.64 & 45.89 & 45.99 \\
  & + Copy & \textbf{54.10} & \textbf{54.40} & \textbf{47.96} & \textbf{48.27} & \textbf{43.27} & \textbf{43.60} & \textbf{39.17} & \textbf{39.51} & \textbf{64.60} & \textbf{64.69} & \textbf{46.09} & \textbf{46.17} \\
  \hline\hline
  \multirow{8}{*}{\raisebox{\normalbaselineskip}[0pt][0pt]{\rotatebox[origin=c]{90}{\textbf{\footnotesize{Esperanto}}}}} 
  & KN & 18.12 & 17.8 & 6.91 & 6.64 & 4.18 & 4.0 & 2.9 & 2.79 & 37.48 & 36.9 & 31.05 & 30.74 \\
  & KN\textsubscript{ext} & 25.17 & 24.93 & 16.44 & 16.3 & 11.99 & 11.92 & 8.77 & 8.79 & 44.93 & 44.77 & 33.77 & 33.71 \\
  & IR & 43.01 & 42.61 & 33.67 & 33.46 & 28.16 & 28.07 & 24.35 & 24.3 & 46.75 & 45.92 & 20.71 & 20.46 \\
  & IR\textsubscript{ext} & \textbf{52.75} & \textbf{51.66} & 43.57 & 42.53 & 37.53 & 36.54 & 33.35 & 32.41 & 58.15 & 57.62 & 31.21 & 31.04 \\ 
  & Ours & 49.34 & 49.40 & 42.83 & 42.95 & 38.28 & 38.45 & 34.66 & 34.85 & 66.43 & \textbf{67.02} & 40.62 & \textbf{41.13} \\
  & + Copy & 50.22 & 49.81 & \textbf{43.57} & \textbf{43.19} & \textbf{38.93} & \textbf{38.62} & \textbf{35.27} & \textbf{34.95} & \textbf{66.73} & 66.61 & \textbf{40.80} & 40.74 \\
  \hline
\end{tabular}
\caption{Automatic evaluation of our model against all other baselines using BLEU 1-4, ROUGE and METEOR for both Arabic and Esperanto Validation and Test set.}
\label{table:automatic-evaluation}
\end{table*}

\vspace{0.1cm} \noindent \textbf{KN} is a 5-gram Kneser-Ney (KN)~\cite{Heafield_13} language model. KN has been used before as a baseline for text generation from structured data \cite{lebret2016} and provided competitive results on a single domain in English.
We also introduce a second KN model (KN\textsubscript{ext}), which is trained on summaries with the special tokens for copy actions. 
During test time, we use beam search of size $10$ to sample from the learned language model.

\vspace{0.1cm} \noindent \textbf{IR} is an Information Retrieval (IR) baseline similar to those that have been used in other text generative tasks~\cite{RushCW15,du_2017}.
First, the baseline encodes the list of input triples using TF-IDF followed by LSA~\cite{halko_svd_2011}.
For each item in the test set, we perform K-nearest neighbors to retrieve the vector from the training set that is the closest to this item and output its corresponding summary.
Similar to KN baseline, we provide two versions of this baseline IR and IR\textsubscript{ext}. 
\begin{figure}[t]
  \centering
  \includegraphics[width=1.0\linewidth]{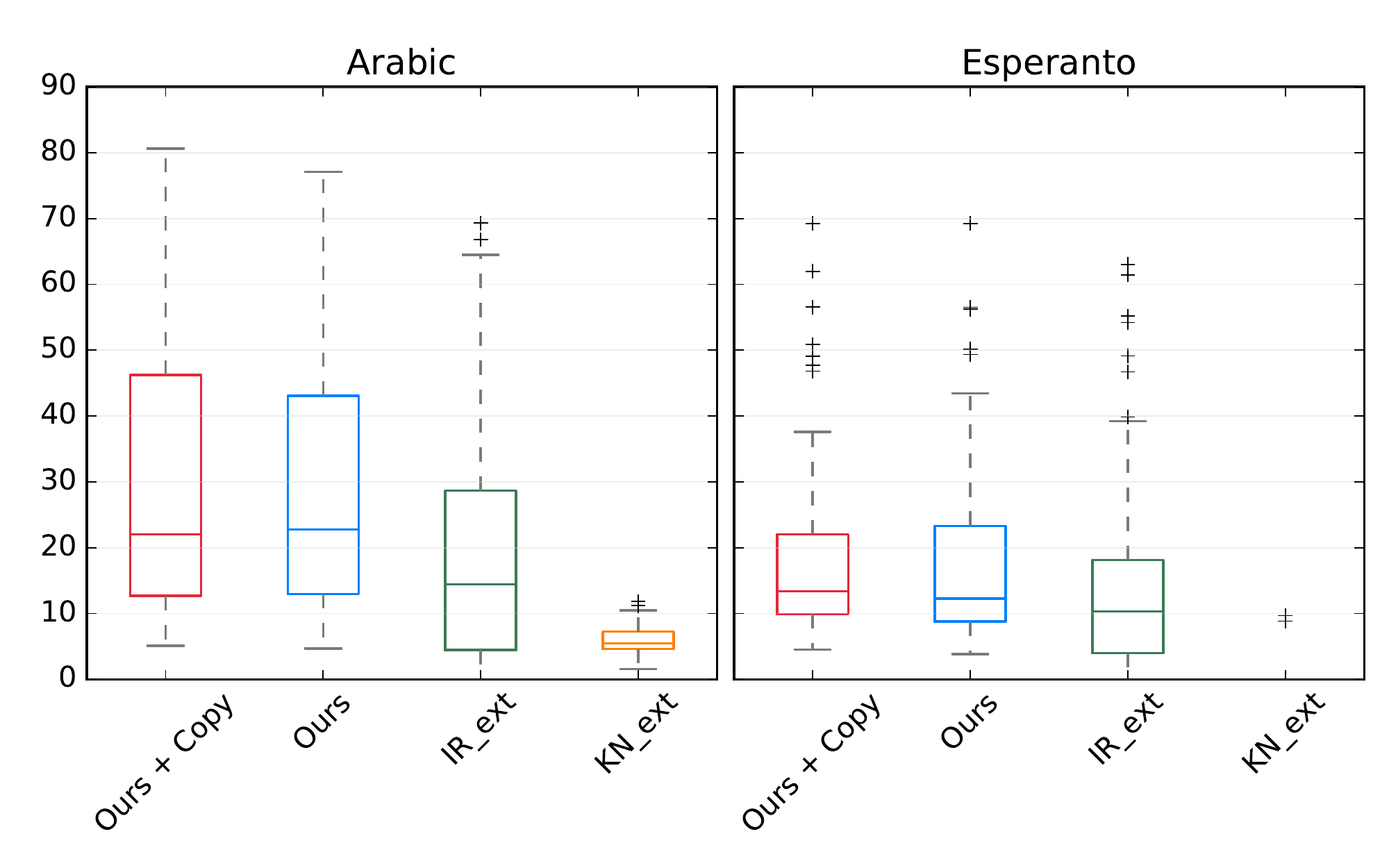}
  \caption{A box plot showing the distribution of BLEU $4$ scores of all systems for each category of generated summaries.}
  \label{fig:categorybleu}
\end{figure}

\section{Results and Discussion}
We evaluate the generated summaries from our model and each of the baselines against their original counterparts from Wikipedia.
Triples sets whose generated summaries are incomplete\footnote{Around $\leq 1\%$ and $2\%$ of the input validation and test triples sets in Arabic and Esperanto respectively led to the generation of summaries without the \texttt{<end>} token. We believe that this difference is explained by the limited size of the Esperanto dataset that increases the level of difficulty that the trained models (i.e. with or without Copy Actions) to generalise on unseen data.} (i.e. summaries for which the special end-of-summary \texttt{<end>} token is generated) are excluded from the evaluation.
We use a set of evaluation metrics for text generation: BLEU~\cite{bleu}, METEOR~\cite{meteor} and ROUGE\textsubscript{L}~\cite{lin2004rouge}. As displayed in Table~\ref{table:automatic-evaluation}, our model shows a significant enhancement compared to our baselines across the majority of the evaluation metrics in both languages. We achieve at least an enhancement of at least $5.25$ and $1.31$ BLEU 4 score in Arabic and Esperanto respectively over the IR\textsubscript{ext}, the strongest baseline.
The introduction of the copy actions to our encoder-decoder architecture enhances our performance further by $0.61-1.10$ BLEU (using BLEU 4).
In general, our copy actions mechanism benefits the performance of all the competitive systems.
\paragraph{Generalisation Across Domains.}
To investigate how well different models can generalise across multiple domains, we categorise each generated summary into one of $50$ categories according to its main entity instance type (e.g. village, company, football player).
We examine the distribution of BLEU-$4$ scores per category to measure how well the model generalises across domains (Figure~\ref{fig:categorybleu}).
We show that i) the high performance of our system is not skewed towards some domains at the expense of others, and that ii) our model has a good generalisation across domains -- better than any other baseline.
Despite the fact that the Kneser-Ney template-based baseline (KN\textsubscript{ext}) has exhibited competitive performance in a single-domain context \cite{lebret2016}, it is failing to generalise in our multi-domain text generation scenario.

\section{Conclusions}
In this paper, we show that with the adaptation of the encoder-decoder neural network architecture for the generation of summaries we are able to overcome the challenges introduced by working with underserved languages.
This is achieved by leveraging data from a structured knowledge base and careful data preparation in a multilingual fashion, which are of the utmost practical interest for our under-resourced task, that would have otherwise required a substantial additional amount of data.
Our model was able to perform and generalise across domains better than a set of strong baselines. 

\section*{Acknowledgements}
This research is partially supported by the Answering Questions using Web Data (WDAqua) project, a Marie Sk\l odowska-Curie Innovative Training Network under grant agreement No $642795$, part of the Horizon $2020$ programme.

\bibliography{Bibliography}
\bibliographystyle{acl_natbib}

\end{document}